\pdfoutput=1
\documentclass[11pt]{article}

\usepackage[preprint]{acl}

\usepackage{times}
\usepackage{latexsym}

\usepackage[T1]{fontenc}
\usepackage[utf8]{inputenc}

\usepackage{microtype}

\usepackage{inconsolata}

\usepackage{graphicx}

\usepackage{booktabs}       % professional-quality tables

\usepackage{geometry}
\usepackage{listings}
\usepackage{xcolor}

\usepackage{caption}
\usepackage{subcaption}

\usepackage{amsmath,amssymb,amsfonts}

\lstdefinelanguage{MyLang}{
    morekeywords={Context, Query, Answer, Prompt, Rule},
    sensitive=false,
    morecomment=[l]{//},
    morestring=[b]"
}

\title{Aligning Model Evaluations with Human Preferences: Mitigating Token Count Bias in Language Model Assessments}

\author{Roland Daynauth \\
  University of Michigan \\
  \texttt{daynauth@umich.edu} \\\And
  Jason Mars \\
  University of Michigan \\
  \texttt{profmars@umich.edu} \\}

\begin{document}
\maketitle
\begin{abstract}
The SLAM paper demonstrated that on-device Small Language Models (SLMs) are a viable and cost-effective alternative to API-based Large Language Models (LLMs), such as OpenAI's GPT-4, offering comparable performance and stability. However, SLAM also identified discrepancies between human preferences and traditional auto-evaluators. This follow-up paper explores methods to align LLM evaluator preferences with human evaluations by addressing biases, particularly toward higher token counts. We employed Bayesian statistics and a t-test to quantify this bias and developed a recalibration procedure to adjust the GPTScorer. Our findings significantly improve aligning the recalibrated LLM evaluator with human evaluations across multiple use cases. For instance, spearman's ranking correlation score in the Recommendation use case improved from -27.27 to 44.55. These results highlight the importance of accounting for biases in automated evaluations to ensure fair and accurate model assessments. The recalibration process enhances the reliability of automated evaluators, leading to better AI models that align with human values and expectations. This study provides a robust methodology for future research into bias correction and emphasizes the feasibility and benefits of developing human-aligned AI evaluation systems.
\end{abstract}

\section{Introduction}

\begin{figure}[t]
\centering
\includegraphics[width=0.9\columnwidth]{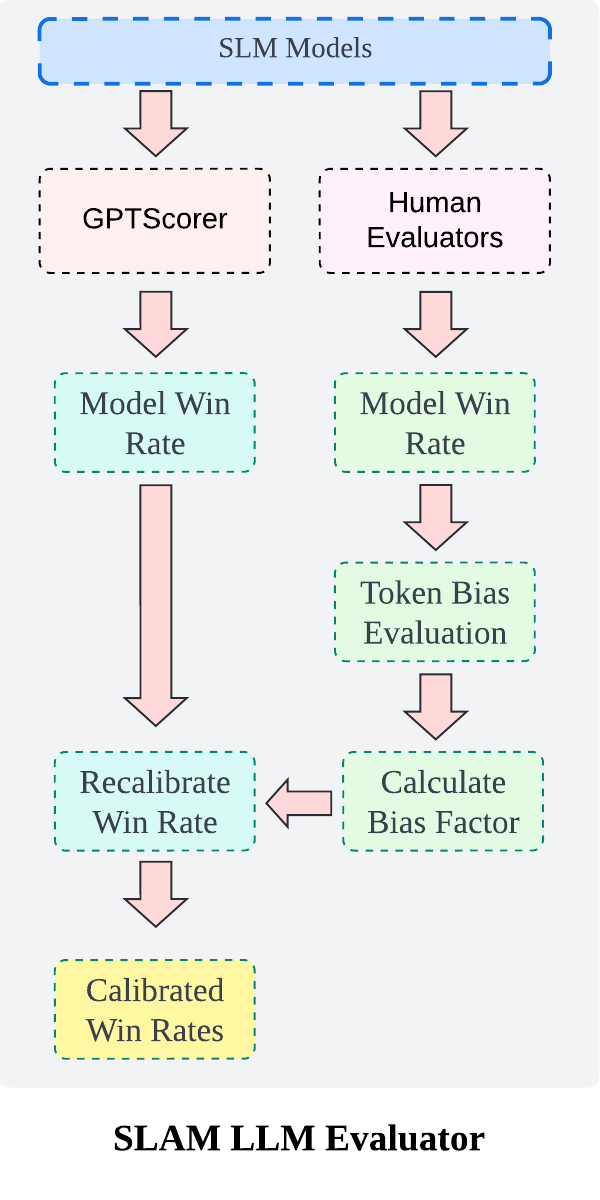}
\caption {Architecture of the SLAM LLM Evaluator}
\label{fig:slm_rank}
\end{figure}

\subsection{Background}
The rapid advancement of Large Language Models (LLMs), such as OpenAI's GPT-4~\cite{openai_chatgpt}, has revolutionized natural language processing tasks across various applications. These models have demonstrated exceptional capabilities in generating human-like text, understanding context, and performing multiple language tasks. However, deploying these models often involves significant costs and dependency on external APIs, which can be prohibitive for many organizations.

In contrast, on-device Small Language Models (SLMs) present a cost-effective and stable alternative\cite{touvron2023llama, jiang2023mistral, neuralchat, openllms23, mitra2023orca, stablelm, starling2023, vicuna2023, tunstall2023zephyr}. SLMs can be deployed locally, reducing the reliance on external services and offering comparable performance for specific use cases. The SLAM paper explored this potential, demonstrating that SLMs could be viable substitutes for API-based LLMs in production environments \cite{slam1}. Despite these promising results, SLAM highlighted a critical issue: the misalignment between human preferences and the evaluations provided by traditional automated evaluators.

Human evaluators often have different criteria and biases, such as a preference for outputs with higher token counts, which are not adequately captured by existing auto-evaluation metrics like SBERT and BERTScore \cite{reimers2019sentence, zhang2020bertscore}. This discrepancy challenges ensuring that model performance evaluations are fair and reflect human judgments. The need to bridge this gap between human preferences and automated assessment is paramount for the reliable deployment of AI models.

\subsection{Objectives}
The primary objective of this work is to address the misalignment between LLM evaluator preferences and human evaluations by identifying and correcting biases inherent in automated evaluation metrics. Specifically, we hypothesize that human evaluators are biased toward selecting models with higher token counts, which may distort the perceived performance of different models. To investigate this, we employ Bayesian statistics and statistical testing to measure and quantify the bias in human evaluations.

Our approach involves the following key steps:

\begin{itemize}
    \item \textbf{Identify and Quantify Bias}: Utilize Bayesian statistics to compare the win probabilities of models with and without higher token counts and perform statistical tests to confirm the presence of bias.
    \item \textbf{Recalibrate Evaluator}: Develop a recalibration procedure to adjust the GPTScorer, an LLM evaluator, to account for the identified bias, thereby improving its alignment with human preferences.
    \item \textbf{Evaluate and Validate}: Conduct a comprehensive analysis of human evaluation data across four specific use cases: All Task (AT), First Task (FT), Peptalk (PT), and Recommendation (RE). Compare the original and recalibrated GPTScorer's correlation with human rankings to demonstrate the effectiveness of our recalibration approach.
\end{itemize}

By achieving these objectives, we seek to enhance the practical applicability of SLMs in production environments, ensuring that these models perform well and align closely with user expectations and preferences.

\section{SLAM Evaluation Methodology}

We conduct a case study of a real AI feature in a production application with four distinct use cases. This study investigates the alignment between human evaluations and automated evaluators in Small Language Models (SLMs) versus Large Language Models (LLMs). First, we describe the application and its key AI features. Subsequently, we present human and automated evaluations for each use case, demonstrating that traditional auto evaluators, such as SBERT and BERTScore, significantly diverge from human preferences. This misalignment underscores the necessity of recalibrating automated evaluators to reflect human judgments better.

\subsection{Use-Cases Introduction}
The application in this case study is Myca, a personal task-management and productivity application.
In Myca, users create and manage their plans and tasks across all aspects of their lives, such as work, personal health, and finances, to stay organized, focused, and productive.
In addition, users can set their longer-term goals and daily habits. Myca records the user's daily accomplishments, progress toward goals, habits, etc.

In this paper, we focus on five of its feedback features that help its users improve their productivity: 

\begin{itemize}
    \item \textit{First Task (FT)}: The "First Task" (FT) message is a personalized note provided to users when they create tasks on the Myca platform. It aims to commend and encourage users to take proactive steps in managing their responsibilities and offers motivation and support as they begin their journey with Myca.
    \item \textit{All Task Completed (AT)}: The "All Task Completed" (AT) message is a commendatory note given to the user after they complete all their focused tasks for the day. It provides a high-level summary highlighting their accomplishments and offers encouragement and motivation in a natural, flowing narrative.
    % \item \textit{New User (NU)}: New application users are given a welcome message that introduces them to the platform's core philosophies and mission, reflecting a commitment to helping users find purpose, achieve personal growth, and maintain mental well-being. It also highlights Myca's focus on productivity and life enjoyment, setting a positive tone for new users.
    \item \textit{Daily Pep Talk (PT)}: At the beginning of every day, an encouraging message is presented to the user based on what they have accomplished the day before, what they plan to do today, and their progress toward their goals.
    \item \textit{Recommendation (RE)}: The "Recommendation" (PT) message is a personalized suggestion provided to users on their first day using Myca, helping them organize their tasks into manageable categories. This message leverages an understanding of the created tasks to propose relevant categories to enhance the user's productivity and organization. The recommendation is concise and avoids listing tasks individually, instead focusing on justifying the suggested categories.
\end{itemize}

\subsection{Human Evaluation}
During the evaluation process, each human evaluator is presented with ten evaluation problems, each from one of four application use-case scenarios, the original input prompt, and two candidate responses generated by the SLMs.
The human evaluator is asked to choose their preferred response based on their judgment of its quality and relevance to the original problem and intention described in the input prompt. If the user is satisfied with neither response nor is deemed equal, they may choose "About the Same" as their rating.

To prevent bias, a blind test is administered, in which the model that generated the response is not disclosed to the evaluator. This ensures ratings are based solely on the response quality, not preconceived notions about a model's capabilities or popularity. Responses are presented in randomized order, which further helps reduce evaluation bias. 
Each human evaluator must finish the set (all ten) of scorings assigned to them for their response to qualify. This is designed mainly for crowdsourcing to avoid inattentive and incomplete inputs from the crowd workers. These incomplete responses are removed from the final results as part of the aggregation and sanitation stage of the evaluation pipeline. 

\subsubsection{Ranking Human Evaluation}
Evaluating SLMs involves ranking models based on their win rates derived from human evaluations. This section outlines how to calculate win rates and determine the models' ranks. 

The win rate of a model \( i \), denoted as \( A_i \), is defined as the proportion of pairwise comparisons in which model \( i \) is selected as the winner, that is, the win rate $A_i$ is calculated as:

\[
A_i = \frac{W_i}{N_i}
\]

where $W_i$ is the number of games won by model $i$ and $N_i$ is the number of games played by $i$.

In cases where two or more models have identical win rates, additional criteria may be used to break ties. Possible tie-breaking methods include considering the average score of the model outputs (if available) or the total number of comparisons involving the model. However, for simplicity, this study primarily focuses on win rates for ranking.

Let \( R_i \) denote the rank of model \( i \). The model with the highest win-rate \( A_{\max} \) will have \( R = 1 \), the second highest win-rate will have \( R = 2 \), and so forth.

This systematic approach accurately determines each model's win rates and subsequent ranks. This ranking methodology ensures a fair and consistent evaluation of model performance based on human judgments. The resulting ranks provide a transparent and interpretable measure of each model's relative performance.

\subsection{SLAM Auto-Evaluators}
\subsubsection{Traditional Similarity Evaluators}
In the original paper, we utilized traditional evaluators such as SBERT and USE-QA to assess the quality of model outputs. We expanded our evaluation toolkit by including transformer-based metrics such as BERTScore and BLUERTScore. Including these new scores aims to provide a more comprehensive analysis and improve alignment with human preferences.

Furthermore, we use the output from the latest version of ChatGPT-4 to generate the reference output. Specifically, we provide ChatGPT-4 with the same prompts as the LLM Evaluator and use its responses as the reference answers. The automated evaluators then score each SLM output by calculating how similar their answers are to these reference answers. Each SLM is given a score based on the evaluator's metric, and the models are ranked by their respective scores for each evaluator.

\begin{itemize}
    \item \textbf{SBERT}: Sentence-BERT (SBERT) is a modification of the BERT network that uses Siamese and triplet network structures to derive semantically meaningful sentence embeddings that can be compared using cosine similarity \cite{reimers2019sentence}. This approach has been effective in capturing semantic similarity between sentences.
    
    \item \textbf{BERTScore}: BERTScore evaluates text generation by aligning BERT embeddings of the candidate and reference sentences \cite{zhang2020bertscore}. It calculates the precision, recall, and F1 score based on the contextual embeddings from BERT, providing a robust semantic similarity measure.
    
    \item \textbf{USE}: The Universal Sentence Encoder (USE) encodes sentences into high-dimensional vectors that capture their semantic content. These vectors can compute cosine similarity scores, offering a measure of semantic equivalence between sentences \cite{cer2018universal}. USE is included in SLAM2 to enhance the evaluation's ability to capture diverse semantic nuances.
    
    \item \textbf{TF-IDF}: Term Frequency-Inverse Document Frequency (TF-IDF) is a traditional information retrieval metric that assesses the importance of words in a document relative to a corpus. It compares the relevance of terms in model outputs to reference texts, providing a syntactic level of evaluation \cite{rajaraman2011mining}. Including TF-IDF in SLAM2 helps balance semantic evaluations with syntactic relevance.
    
    \item \textbf{BLEURTScore}: BLEURT (Bilingual Evaluation Understudy with Representations from Transformers) leverages pre-trained transformers and a large amount of synthetic data to predict human judgments of text generation quality \cite{sellam2020bleurt}. BLEURTScore is introduced in SLAM2 to provide a sophisticated measure that combines the strengths of both traditional metrics and deep learning models.
\end{itemize}

\subsubsection{GPTScorer}
GPTScorer is an evaluation tool designed to assess the performance of language models by comparing their generated outputs to reference outputs. It leverages the capabilities of GPT-4 to provide detailed and nuanced evaluations. GPTScorer operates by judging the best model based on its use case. Like the human evaluation, GPTScorer is given a prompt from a randomly selected use case and response outputs from two SLMs. It is prompted to choose the base response for that particular use case and provide a reason for its selection. The prompt for GPTScorer is shown below.

\begin{lstlisting}[language=MyLang, caption={GPTScorer Prompt}]
[Use case]
{usecase}

[Response A]
{response_a}

[Response B]
{response_b}

For the given use case, out of Response 
A and Response B, Select the most 
suitable output for the given usecase. 
Provide Reasoning for the choice and 
Output as a dictionary with a key 
"overall". valueout of("Response A", 
"Response B", "About the Same"). 
Follow the following template.

[Reasoning] <Reason>
[Output] <Output>

\end{lstlisting}

\subsection{Experimental Result}

\begin{table}
\small
\begin{tabular}{lrrrr}
\toprule
 & AT & FT &  PT & RE \\
\midrule
USE & -30.91 & 39.09 & 13.64 & -43.64 \\
Tfidf & 1.82 & 7.27  & 1.82 & \textbf{28.18} \\
Sbert & -29.09 & -4.55 &  0.91& -36.36 \\
BertScore & \textbf{30.91} & \textbf{53.64} &  -30.91 & -31.82 \\
BleurtScore & -23.64 & 18.18  & 24.55 & -44.55 \\
GPTScorer & 25.45 & 7.27  & \textbf{43.64} & \textit{-27.27} \\

\bottomrule
\end{tabular}
\caption{Ranking Correlation score between Human Rankings and Auto Evaluators}
\label{tab:eval_results}
\end{table}

Table ~\ref{tab:eval_results} presents the ranking correlation scores between human rankings and various automated evaluation methods. The evaluation results reveal varied alignment between automatic and human evaluators across five use cases. These results suggest that while BERTScore and GPTScore excel in specific scenarios, no evaluator consistently matches human preferences across all use cases, underscoring the need for tailored metrics.

\section{Measuring human token count bias in pairwise evaluation}
In exploring the alignment of LLM evaluator preferences with human evaluations, we suspected that humans may inherently favor outputs with higher token counts when evaluating SLMs. To investigate this potential bias, we employed Bayesian statistics to measure and quantify the extent of this preference.

\begin{figure}[h]
    \centering
    \begin{subfigure}[b]{0.4\textwidth}
        \centering
        \includegraphics[width=\textwidth]{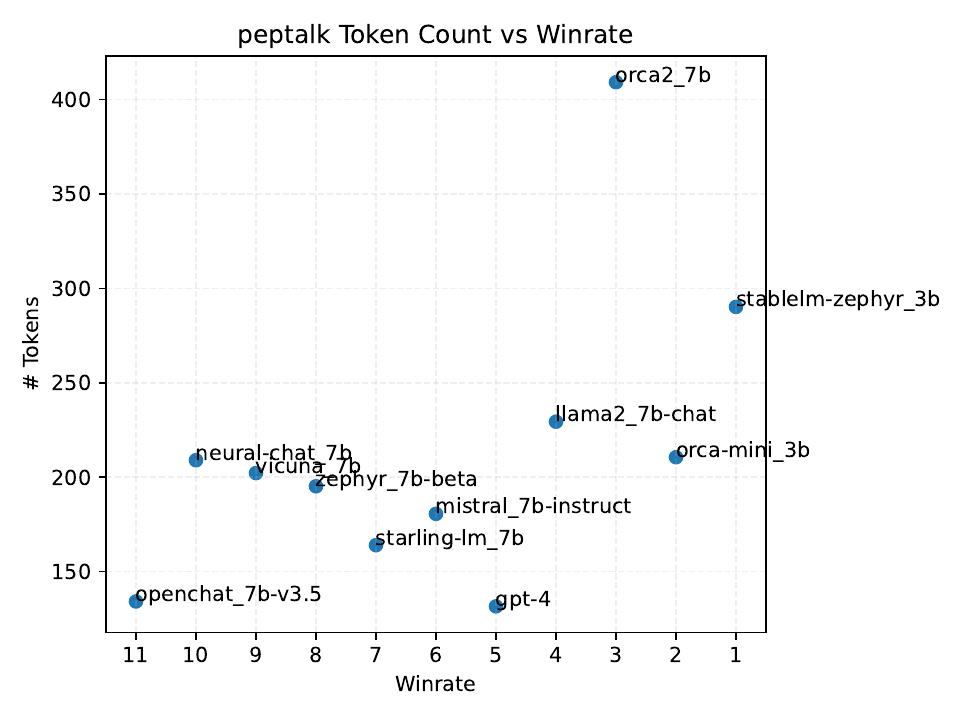} % Replace with your image file
        \caption{Token Count vs. human evaluation win rate for the peptalk use case}
        \label{fig:token1}
    \end{subfigure}
    \hfill
    \begin{subfigure}[b]{0.4\textwidth}
        \centering
        \includegraphics[width=\textwidth]{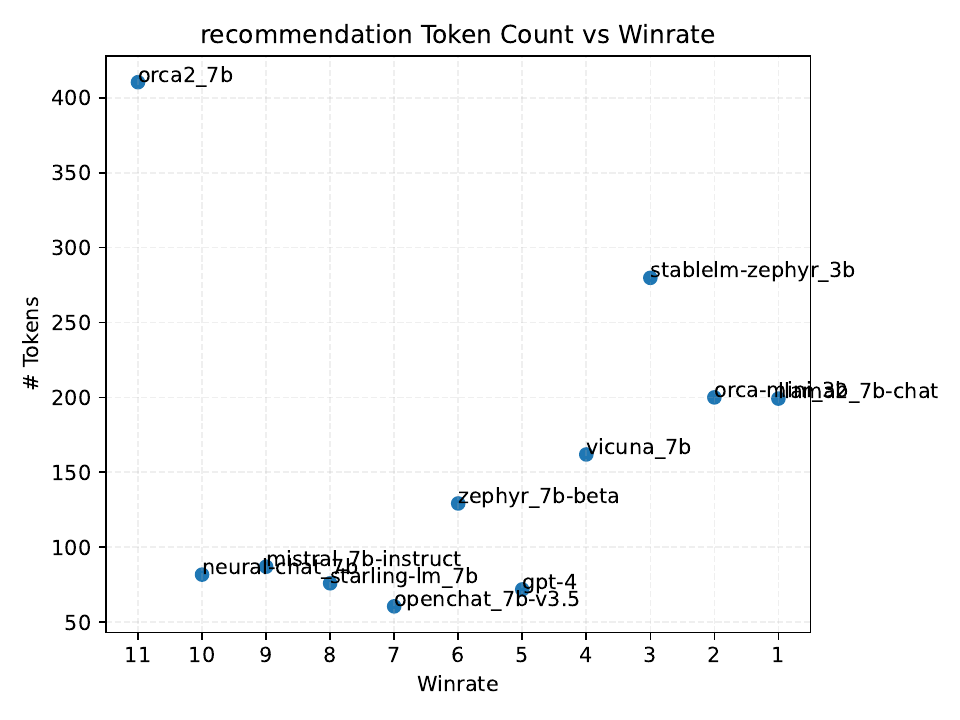} % Replace with your image file
        \caption{Token Count vs. human evaluation win rate for the peptalk use case}
        \label{fig:token2}
    \end{subfigure}
    \caption{Scatter plots showing the alignment between human SLM preference and the length of their output}
    \label{fig:tokens}
\end{figure}

Our starting point was the belief that when presented with two outputs from different models, human reviewers are more likely to select the one with a higher token count as the winner. As shown in Figure \ref{fig:tokens}, a scatter plot reveals a positive correlation between the human rankings of the models and the number of tokens in their outputs. This positive correlation suggests that human evaluators may indeed be influenced by the length of the output, favoring models that produce more extended responses.

If this bias is present, it could skew the evaluation results and misalign automated evaluators with human preferences. Understanding and addressing this bias is crucial to ensure that the evaluation process fairly and accurately reflects the actual performance of the models without undue influence from the length of their outputs.

\subsection{Measuring Token Bias with Bayesian Statistics}
To measure this bias, we utilized Bayes' Theorem, a foundational principle in probability theory that allows us to update our beliefs based on new evidence. In this context, Bayes' Theorem helps us calculate the probability that a model wins, given that it has a higher token count than its competitor.

Bayes' Theorem is expressed as follows:

\[
P(A_i \mid B_i) = \frac{P(B_i \mid A_i) \cdot P(A_i)}{P(B_i)}
\]

Where:

\begin{itemize}
    \item \( P(A_i \mid B_i) \) is the probability that model \( i \) wins given that it has a higher token count.
    \item \( P(B_i \mid A_i) \) is the probability that model \( i \) has a higher token count given that it wins.
    \item \( P(A_i) \) is the prior probability that model \( i \) wins.
    \item \( P(B_i) \) is the overall probability that model \( i \) has a higher token count.
\end{itemize}

To assess the bias, we need to determine if \( P(A_i) \) is significantly higher when \( B_i \) is high. Specifically, we compare \( P(A_i \mid B_i > t) \), where \( t \) is a threshold indicating a higher token count. If \( P(A_i \mid B_i > t) \) is significantly higher than \( P(A_i) \), it suggests a bias towards higher token counts.

\subsubsection{Statistical test for evaluation token bias }
To determine whether human reviewers are biased towards selecting models with higher token counts, we need to test the null hypothesis (\( H_0 \)) that there is no bias. This can be done by comparing the probabilities \( P(A_i) \) (the probability that model \( i \) wins) and \( P(A_i \mid B_i > t) \) (the probability that model \( i \) wins given it has a higher token count). Here, we explain the statistical tests used to determine whether the null hypothesis is accepted or rejected.

To rigorously test this hypothesis, we use the match outcome from the human evaluation dataset. We calculated the overall win probability \( P(A_i) \) as the proportion of matches won by model \( i \) out of the total matches \( N \), and the conditional win probability \( P(A_i \mid B_i > t) \) as the proportion of matches won by model \( i \) out of those where it had a higher token count \( T \). We then computed the variances and standard deviations for these probabilities.

The t-test was used to compare the means of these win probabilities. The t-statistic was calculated as follows:

\[
t = \frac{\bar{X}_1 - \bar{X}_2}{\sqrt{\frac{s_1^2}{N} + \frac{s_2^2}{T}}}
\]

where \( \bar{X}_1 \) and \( \bar{X}_2 \) are the mean win probabilities, and \( s_1 \) and \( s_2 \) are the standard deviations. The degrees of freedom for the test were determined using:

\[
df = \frac{\left( \frac{s_1^2}{N} + \frac{s_2^2}{T} \right)^2}{\frac{\left( \frac{s_1^2}{N} \right)^2}{N - 1} + \frac{\left( \frac{s_2^2}{T} \right)^2}{T - 1}}
\]

We calculated the p-value from the t-distribution with these degrees of freedom. If the p-value was less than the significance level (e.g., \( \alpha = 0.05 \)), we rejected the null hypothesis (\( H_0 \)) that there is no bias, concluding that human reviewers do show a bias towards selecting models with higher token counts. This statistical approach allowed us to quantify and validate the hypothesized bias, ensuring our evaluations align more closely with human preferences.

\subsection{Calibrating LLM Evaluation for Token Bias}
In our effort to align Large Language Model (LLM) evaluator preferences with human evaluations, we identified a potential bias in human reviewers favoring models with higher token counts. To address this bias and ensure fairer evaluations, we developed a procedure to recalibrate LLM evaluations, making them more consistent with human preferences.

Using Bayes' Theorem, we calculated the conditional probability that a model wins given that it has a higher token count (\( P(A_i \mid B_i) \)), and compared this to the probability that a model has a higher token count given that it wins (\( P(B_i \mid A_i) \)). Specifically, the ratio \( \frac{P(A_i \mid B_i)}{P(B_i \mid A_i)} \) simplifies to \( \frac{P(A_i)}{P(B_i)} \), representing the relative likelihood of winning versus having a higher token count. This ratio indicates bias, with significant deviations suggesting an over- or under-influence of token count on win probabilities.

We introduced an adjustment factor to account for the bias. The factor was calculated based on the ratio of the human win-rate to the LLM win-rate and the observed token count probability:
   \[
   \beta_i = \frac{P(A_i \mid B_i)}{P(B_i \mid A_i)} = \frac{P(A_i)}{P(B_i)}
   \]

Using the adjustment factor, we re-calibrated the LLM win rates:
   \[
   P(A_i^{Adjusted}) = P(A_i^{LLM}) \cdot \beta_i
   \]

% $\lambda_i$ > 1 suggests that winning is more likely than having a higher token count, implying that token count might not be a significant factor in winning. In contrast, $\lambda_i < 1$ suggests that having a higher token count is more likely than winning, implying that a higher token count might not translate directly to higher chances of winning or indicating a potential underappreciation of the token count by the human reviewers.

% \[
% \frac{P^{\prime}(A_i)}{P^{\prime}(B_i)} = \lambda^{\prime}_i \implies P^{\prime}(A_i) = P^{\prime}(B_i)\lambda^{\prime}_i
% \]

\section{Results and Discussion}
In this section, we present and discuss the findings of our study on recalibrating LLM evaluations to align more closely with human preferences. The primary focus is on the impact of recalibrating the GPTScorer to account for identified token count bias, thereby improving its correlation with human rankings. The results are summarized in Table \ref{tab:recal_result}, which compares the correlation scores between human rankings and the original and recalibrated GPTScorer.

\subsection{Results}
Table \ref{tab:recal_result} shows the ranking correlation scores across four use cases: All Task (AT), First Task (FT), Peptalk (PT), and Recommendation (RE). The scores represent the correlation between human rankings and the evaluations provided by the GPTScorer before and after recalibration.

The recalibrated GPTScorer demonstrates a moderate to significant improvement in alignment with human evaluations across all use cases.

\begin{table}
\small
\begin{tabular}{lrrrrr}
\toprule
 & AT & FT &  PT & RE \\
\midrule
GPTScorer & 25.45 & 7.27  & 43.64 & -27.27 \\
 Recalibrated GPTScorer& \textbf{44.55} & \textbf{20.00}  & \textbf{44.55} & \textbf{44.55} \\
\bottomrule
\end{tabular}
\caption{Ranking Correlation score between Human Rankings and the Recalibrated GPTScorer and original GPTScorer showing moderate to significant improvement in alignment between human and LLM evaluators.}
\label{tab:recal_result}
\end{table}

\subsection{Discussion}

The results indicate that recalibrating the GPTScorer to account for token count bias improves alignment with human preferences. This improvement is particularly notable in the Recommendation (RE) use case, where the correlation increased from a negative value to a significant positive alignment. This suggests that the original GPTScorer may have been substantially biased by token count, affecting its ability to align with human judgments.

The improvements in the All Task (AT) and First Task (FT) use cases, though moderate, are also significant. They highlight the effectiveness of the recalibration process in enhancing the evaluator's sensitivity to human preferences. The slight improvement in the Peptalk (PT) use case suggests that the original GPTScorer was already relatively well-aligned with human evaluations, but recalibration still provided a minor enhancement.

These findings underscore the importance of accounting for biases in automated evaluators. By identifying and correcting for token count bias, we can develop evaluators that provide more accurate and human-aligned assessments of model performance. This improves the reliability of automated evaluations and ensures that the models selected for deployment truly reflect human preferences.

\subsection{Implications for future research}
The methodology and findings from this study provide a foundation for further research into bias correction in LLM evaluators. Future work could explore additional sources of bias and develop more sophisticated recalibration techniques. Additionally, expanding the evaluation to include a broader range of use cases and more diverse sets of human evaluators could provide deeper insights into the generalizability of the recalibration process.

This study demonstrates that recalibrating LLM evaluators to align with human preferences is feasible and beneficial. The improved correlation scores achieved through recalibration highlight the potential for developing more accurate and fair evaluation systems, ultimately leading to better AI models that align with human values and expectations.
\section{Related Work}

Language model evaluation has seen significant advancements with the development of Large Language Models (LLMs) such as OpenAI's GPT-3 and GPT-4. These models have demonstrated exceptional capabilities in various natural language processing tasks \cite{brown2020language}. Despite their success, the reliance on proprietary LLMs presents several challenges, including high costs, performance variability, and lack of control over the models.

Researchers have explored using SLMs as viable alternatives to address these challenges. The SLAM study systematically evaluated the feasibility of using on-device SLMs to replace API-based LLMs. It was shown that SLMs could offer comparable performance, enhanced stability, and significant cost savings \cite{slam1}. However, the study highlighted the gap in understanding human preferences for LLM versus SLM outputs.

Traditional evaluators such as SBERT and BERTScore have been commonly used for automated evaluation of model outputs \cite{reimers2019sentence, zhang2020bertscore}. These methods rely on comparisons to reference responses, often failing to capture the nuances of human preferences. As a result, there has been a growing interest in aligning automated evaluators with human evaluations.

Numerous studies have investigated the alignment of Large Language Models (LLMs) with human preferences, mainly focusing on evaluating LLM performance through pairwise ranking. \cite{10.1145/3460231.3474274} concentrated on debiased explainable pairwise ranking from implicit feedback, highlighting the Bayesian Personalized Ranking model's capability to handle implicit feedback. \cite{10.18653/v1/2022.findings-naacl.147} introduced a method for interactive text summarization with preference feedback, integrating preference learning to simulate human evaluation. Additionally, \cite{10.1609/aaai.v38i17.29865} proposed Preference Ranking Optimization (PRO) as a technique to fine-tune LLMs for human alignment.

This study aims to address these challenges by recalibrating LLM evaluators using statistical analysis of human evaluation data. Initial results have shown promising improvements in alignment, suggesting that better alignment can be achieved without expensive fine-tuning or few-shot evaluations.

\section{Conclusion}

In this study, we addressed the critical issue of aligning large language model (LLM) evaluations with human preferences by identifying and mitigating biases inherent in automated evaluation metrics. Our primary focus was on the bias toward higher token counts, which we hypothesized would influence human evaluators when comparing Small Language Models (SLMs). Through rigorous statistical analysis and recalibration of the GPTScorer, we aimed to enhance the alignment between LLM evaluators and human judgments.

Our findings reveal that the recalibrated GPTScorer showed significant improvements in correlation with human evaluations across multiple use cases. The most substantial improvement was observed in the Recommendation (RE) use case, where the correlation score increased from a negative value to a strong positive alignment. Other use cases, such as All Task (AT) and First Task (FT), also demonstrated moderate improvements, while the Peptalk (PT) use case saw a slight enhancement.

These results underscore the importance of accounting for biases in automated evaluators to ensure fair and accurate assessments of model performance. By implementing a bias adjustment factor based on the ratio \( \frac{P(A \mid B)}{P(B \mid A)} \), we successfully recalibrated the GPTScorer, leading to better alignment with human preferences. This recalibration process improves the reliability of automated evaluations and ensures that the models selected for deployment genuinely reflect human values and expectations.

Our study provides a foundational methodology for future research into bias correction in LLM evaluators. Future work could extend this approach to explore additional biases and develop more sophisticated recalibration techniques. Moreover, expanding the evaluation to encompass a broader range of use cases and a more diverse set of human evaluators would provide deeper insights into the generalizability of our findings.

In conclusion, recalibrating LLM evaluators to align with human preferences is feasible and advantageous. The improved correlation scores achieved through our recalibration process highlight the potential for developing more accurate and fair evaluation systems, ultimately leading to the deployment of better AI models that are aligned with human judgments. This study represents a significant step towards ensuring that AI systems are evaluated and selected to reflect human values and preferences, paving the way for more trustworthy and effective AI technologies.

\bibliography{main}
\appendix

\end{document}